\pdfoutput=1
\PassOptionsToPackage{table}{xcolor}
\documentclass[11pt]{article}

\usepackage[preprint]{acl}

\usepackage{times}
\usepackage{latexsym}
\usepackage{graphicx}

\usepackage[T1]{fontenc}

\usepackage[utf8]{inputenc}

\usepackage{microtype}

\usepackage{inconsolata}

\usepackage{comment}
\usepackage{overpic}

\usepackage{worldflags}

\title{\textit{What's in a prompt?} \\ Language models encode literary style in prompt embeddings \\ ~}

\author{Rapha\"el Sarfati \\
  Cornell University \\
  \texttt{raphael.sarfati@cornell.edu} \\\And
    Haley Moller \\
  Yale University \\
  \texttt{haley.moller@yale.edu} \\ \And
    Toni J. B. Liu \\
  Cornell University \\
  \texttt{jl3499@cornell.edu} \\ \AND
    Nicolas Boull\'e \\
  Imperial College London \\
  \texttt{n.boulle@imperial.ac.uk} \\ \And
    Christopher Earls \\
  Cornell University \\
  \texttt{earls@cornell.edu} \\}

\begin{document}
\maketitle

\begin{abstract}
Large language models use high-dimensional latent spaces to encode and process textual information. 
Much work has investigated how the conceptual content of
words translates into geometrical relationships
between their vector representations.
Fewer studies analyze how the \emph{cumulative} information of an entire prompt becomes condensed into individual embeddings under the action of transformer layers.
We use literary pieces to show that information about intangible,
rather than factual, aspects of the prompt are contained in deep representations.
We observe that short excerpts ($10 - 100$ tokens) from different novels separate in the latent space independently from what next-token prediction they converge towards.
Ensembles from books from the same authors are much more entangled than across authors, suggesting that embeddings encode stylistic features. 
This \emph{geometry of style} may have applications for authorship attribution and literary analysis, but most importantly reveals the sophistication of information processing and compression accomplished by language models.
\end{abstract}

\section{Introduction}

\textit{``What's in a name?''} famously asked Juliet~\citep{Juliet} to interrogate the relationship between a concept's multifaceted reality and its shorthand designation in the form of a word.\footnote{~\textit{``That which we call a rose / By any other word would smell as sweet.'' Act II, Scene II}}

Four hundred years later, the question finds renewed significance in the context of large language models (LLMs)~\cite{brown2020language,touvron2023llama,grattafiori2024llama}, where words are represented as vectors in a high-dimensional latent space~\cite{mikolov2013efficientestimationwordrepresentations,mikolov2013distributed}. 
Much research has attempted to elucidate what information these representations, also called `embeddings', convey, and how this information is encoded.
Some fascinating insights have been uncovered in terms of geometrical relationships between concepts~\citep{mikolov2013efficientestimationwordrepresentations,park2024linearrepresentationhypothesisgeometry,park2025geometrycategoricalhierarchicalconcepts}.

\begin{figure}[htbp] 
\begin{center}
\centerline{\includegraphics[width = \columnwidth]{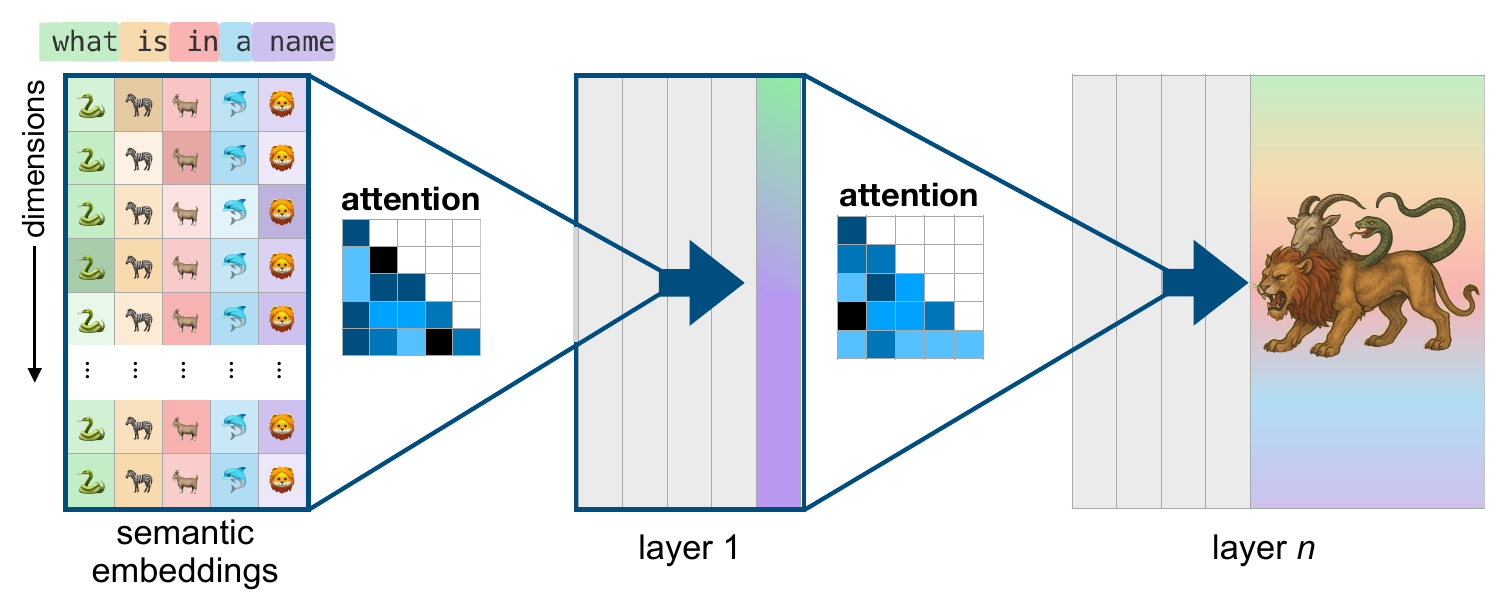}}
\caption{
After semantic embedding of the prompt, vectors represent a single word. 
As the prompt passes through transformer layers, the attention mechanism funnels more and more information about preceding tokens into the last embedding -- turning it into a `chimera' vector, encoding bits of information from all others.
}
\label{fig:chimera}
\vspace{-2em}
\end{center}
\end{figure}

Yet, word-to-vec(tor) embedding is only the first step.
For LLMs, an embedding leaves its starting point and is transported, transformer layer after transformer layer, to a new location that will determine next-token prediction.
In the process, it loses its original identity and starts accumulating information about all preceding tokens -- and the emergent meaning of their sequence (Fig.~\ref{fig:chimera}).

This raises the question: What's in a \emph{prompt}? 
In other words, what kind of information contained in the sequence of words forming the LLM's input finds itself distilled into deeper embeddings?

Prior work (see also Appendix~\ref{app:background}) has shown that embeddings can contain global factual information about, e.g., whether the preceding statement is true or false~\citep{marks2024geometrytruthemergentlinear}, or its relation to space and time~\citep{gurnee2024languagemodelsrepresentspace}.
This is interesting and sensible: factual understanding seems necessary to output compelling prompt continuation.

Here, we find evidence of the presence of more subtle signals. Using short excerpts from various literary works, we show that the embeddings contain implicit information about the origin of the passage and can be classified with high accuracy. 

This study is \emph{not} aimed at merely assessing the performance of LLMs for authorship attribution~\cite{huang2025authorshipattributionerallms}, but rather at showing that implicit prompt features like authorship are encoded in deep embeddings (and not early ones).

\section{Methods}

\paragraph{Overview.}
We base our analysis on ensembles of short excerpts from various literary works and collect the embeddings \emph{of the last (rightmost) token} after each layer of an LLM.
From these vector representations, we apply classifier techniques to evaluate whether excerpts can be linked to their original oeuvres based on a \emph{single}, information-rich embedding.
We investigate in particular the influence of context length $N$ (number of tokens in the input passage) and layer depth $L$ (number of transformer layers that the prompt has crossed).

\paragraph{Sources.} 
We use digital versions of literary works obtained from the Project Gutenberg website (\verb|gutenberg.org|).
We curate a corpus of 19$^\mathrm{th}$ and early 20$^\mathrm{th}$ century anglophone novels for both consistency and diversity of styles, some of them from the same author~(Appendix~\ref{app:dataset}).

\paragraph{Processing.}
A full novel's text is passed through a model's tokenizer.
The sequence of tokens' IDs is then split into chunks of length $N$ tokens, with $N = 8, 16, \dots, 128$ typically.
Importantly, these text chunks do not correspond to any particular syntactic unit and can end with any kind of token (Appendix~\ref{app:dataset}).
The rightmost embeddings $\vec{x}_N(L)$ are collected after each layer~$L$ for each chunk.\footnote{
Indeed, the last embedding of the prompt is the one that learns from all preceding tokens thanks to the causal masked attention mechanism.}

\paragraph{Models.} 
We use a suite of open-source models from \verb|Hugging Face|.
We focus on the 16-layer \verb|Llama-3.2-1B base| model~\citep{meta2024llama3_2} for insight and extend to other models in Appendix~\ref{app:models}.

\paragraph{Classifiers.}
We use standard supervised classifying techniques to investigate the separability of ensembles of high-dimensional vectors:
Support Vector Machine (SVM) probes for binary classification, and Multilayer Perceptron (MLP) probes for multiclass (details in Appendix~\ref{app:dataset}).

\begin{figure}[htbp]
\centering
\begin{overpic}[width = \columnwidth]{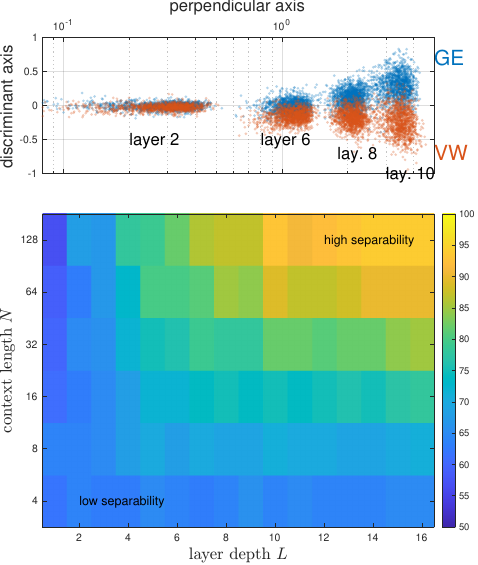}
\put(0,95){\textsf{A}}
\put(0,62){\textsf{B}}
\end{overpic}
\caption{
(A) Ensembles of short excerpts ($N=64$ tokens) from GE and VW separate in the latent space as embeddings travel through successive transformer layers.
(B) Linear classifier accuracy (\%) to distinguish GE vs VW ensembles as a function of prompt's number of tokens $N$ and number of transformer layers crossed~$L$.
}
\label{fig:classifier}
\end{figure}

\section{Results}

\subsection{Embeddings encode authorship}
Does a short passage ($10$-$100$ words) from a novel contain enough information to be properly attributed after processing by an LLM? 
We compare excerpts from two novels: George Eliot's (GE) \textit{Silas Marner} and Virginia Woolf's (VW) \textit{Mrs Dalloway}. 
By training a linear classifier, we examine whether the two ensembles of high-dimensional embeddings in the Llama~3.2~1B model can be separated.
They can.
We observe in Fig.~\ref{fig:classifier} that as their length $N$ increases and the embeddings travel deeper into the model, the excerpts can be classified with over 90\% accuracy.
In contrast, when there is not enough context (small $N$) or not enough attention layers to `cross-pollinate' information across tokens (small $L$), each excerpt's last embedding has not absorbed enough contextual information to reflect authorship.

\begin{figure}[t]
\centering
\begin{overpic}[width = \columnwidth]{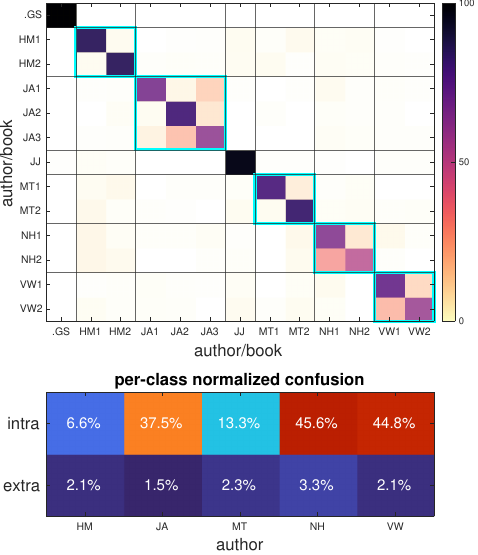}
\put(0,97){\textsf{A}}
\put(0,30){\textsf{B}}
\end{overpic}
\caption{
(A) Accuracy (\%) of an MLP probe to distinguish passages from 13 different books ($N = 128, L = 16$).
See Tab.~\ref{tab:authors} for the list of authors and novels.
Cyan squares emphasize novels from the same authors. 
It is noteworthy that confusion increases between books of the same author, even though they relate to different topics. 
(B) Results specific to probe confusion for books from the same author (intra) or a different author (extra).
}
\label{fig:mlp}
\end{figure}

More generally, an MLP probe can distinguish excerpts from several novels with overall 75\% accuracy (Fig.~\ref{fig:mlp}A). 
These passages represent small snippets of text from various works, with no consistency in theme or syntax~(see Tab.~\ref{tab:excerpts} in Appendix~\ref{app:dataset}).
It could be that they contain enough \emph{factual} information (names, subjects, etc.) to reveal their provenance.
However, we also observe a marked increase in classifier confusion across works from the same authors (Fig.~\ref{fig:mlp}B).
This suggests that the classifier might be relying on patterns of vocabulary and syntax which find themselves encoded in deep embeddings (and not early ones).
We refer to these abstract distinctive features as ``style'' and investigate what exactly is encoded and how.

\subsection{Stylistic signatures align with large principal components}
It's been observed that embeddings and their trajectories tend to form low-dimensional structures.
For example, \citet{viswanathan2025geometrytokensinternalrepresentations} showed that the intrinsic dimension (ID) of token representations from a given prompt is generally much smaller than that of the ambient space.
\citet{sarfati2025linesthoughtlargelanguage} found, using singular value decomposition, that prompt ensembles stretch along a few directions and diffuse about most of the remaining subspace.

\begin{figure}[htbp]
\centering
\begin{overpic}[width=\columnwidth]{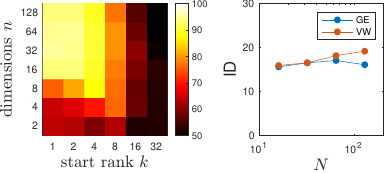}
\put(0,42){\textsf{A}}
\put(57,42){\textsf{B}}
\end{overpic}
\caption{
Dimensionality of stylistic features.
(A) Probe accuracy (\%) for classifying GE vs. VW ensembles projected onto PCA subspaces spanned by $\{\vec{u}_k, \dots, \vec{u}_{k+n-1} \}$, where $\vec{u}_k$ is the $k$-th principal component and $n$ is the subspace dimension
(B) Intrinsic dimension for embedding ensembles as a function of context length $N$.
ID is calculated using the TwoNN method described in~\citet{valeriani2023geometryhiddenrepresentationslarge}. 
}
\label{fig:dimension}
\end{figure}

Is the property of ``style'' contained in the small or large dimensions? Fig.~\ref{fig:dimension} indicates that probe accuracy plateaus at a maximum when keeping about 16 directions along the largest principal components.
Interestingly, however, the ID of the ensembles remains under 20 dimensions and doesn't change with increased context $N$.
This suggests that contextual information effectively moves ensembles into different corners of the latent space rather than altering their shape complexity.

\subsection{Disrupting syntax conserves separability}
Is style inferred from syntactic or semantic features? 
To investigate, we use a shuffling approach introduced in~\citet{viswanathan2025geometrytokensinternalrepresentations}.
For each input pseudo-sentence, blocks of $B$ consecutive tokens are rearranged randomly, with $B = 1$ (every token is independent), $B=4$ (groups of four tokens are kept together), etc.
Perhaps surprising, Tab.~\ref{tab:shuffle} shows that classifier probes remain accurate for all persistence lengths~$B$.
This strongly suggests that the stylistic signature perceived by LLMs might rely more on lexical content than structure.
However, ensembles corresponding to different shuffling scales are perfectly separable, indicating that word order and syntax do affect the representation of a prompt -- as expected.

\begin{table}[htbp]
\centering
\begin{tabular}{l|ccc}
\hline
VW~\textbackslash~GE & $B=1$ & $B=4$  & $B=32$ \\
\hline
$B=1$ & 97 & 100  & 100 \\
$B=4$ & 100 & 95  & 98  \\  
$B=32$ & 100 & 99  & 92  \\ 
\hline
\end{tabular}
\caption{
Linear probe accuracy (\%) for various shuffling block size $B$.
}
\label{tab:shuffle}
\end{table}

\subsection{Geometrical relationships across languages}
Famously, LLMs latent spaces exhibit alluring geometrical properties, notably the parallelogram structures such as \verb|woman:queen::man:king|~\citep{Li2025}.
Do similar relationships exist at the ensemble level?
Tab.~\ref{tab:frvsen} suggests that they do. 
We consider three French novels and their English translations:
Gustave Flaubert's \textit{Madame Bovary}, Victor Hugo's \textit{Ninety-Three}, and \'{E}mile Zola's \textit{Germinal}.
A probe trained to separate a French ensemble pair keeps its accuracy on the corresponding English pair.
Similarly, there is a strong similarity between centroid separations in French and in English. 
The cosine distance between author A and author B in French and in English is between 0.5 and 0.6, which is far smaller than expected for random vectors ($1 \pm 1/\sqrt{2048}$).
This observation seems to generalize point-based geometrical structures to distributed clouds of embeddings.

\begin{table}[htbp]
\centering
\begin{tabular}{l|ccc}
\hline
& \verb|F/H| & \verb|F/Z|  & \verb|H/Z| \\
\hline
\worldflag[width=12pt, length=16pt]{FR} & \cellcolor{violet} \textcolor{white}{79\%} & 63\%  & 65\% \\
\hline
\worldflag[width=12pt, length=16pt]{GB} & 82\% & 65\%  & 63\%  \\  
\hline
\end{tabular}
\caption{
Accuracy of a reference linear probe trained to distinguish Flaubert (F) from Hugo (H) in French, when applied to other ensemble pairs.
The probe achieves about the same accuracy when applied to the corresponding English-translated ensembles. 
It performs significantly worse (60\%) when applied to unrelated pairs involving Zola (Z).
}
\label{tab:frvsen}
\end{table}

\section{Discussion and future directions}
\paragraph{Practicalities.}

We remark that a by-product of training LLMs is that they inherit a fine perception of stylistic and informational patterns, even from short passages.
Perhaps literature scholars will build upon this idea to implement more sophisticated methods to address some long-standing mysteries and controversies:
Was Shakespeare a single writer?
Could \'Emile Ajar have been identified as Romain Gary before illegitimately snatching a second Prix Goncourt~\citep{tirvengadum1996linguistic}?

\begin{figure}[htbp]
\begin{center}
\centerline{\includegraphics[width=\columnwidth]{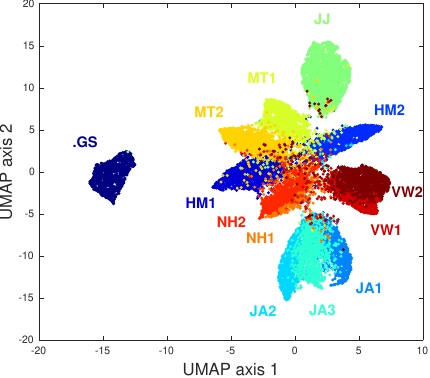}}
\caption{
Map of style: low-dimensional visualization of the high-dimensional geometry across books and authors.
Text chunks ($N=128, L=16$) are UMAP embedded from their 32-dimensional activations extracted at the penultimate layer of the MLP classifier of Fig.~\ref{fig:classifier}.
We note the substantial overlap between excerpts from the same author, e.g., Austen (\textcolor[rgb]{0.26,0.5,0.97}{ \small \bf \textsf{JA1}},
\textcolor[rgb]{0.4,0.8,1}{ \small \bf  \textsf{JA2}},
\textcolor[rgb]{0.5,1,0.8}{ \small \bf  \textsf{JA3}}) 
or Wolf 
(\textcolor[rgb]{0.75,0.17,0.09}{ \small \bf \textsf{VW1}},
\textcolor[rgb]{0.45,0.09,0.04}{ \small \bf \textsf{VW2}}).
More comments in Appendix~\ref{app:analysis}.
}
\label{fig:map}
\vspace{-2em}
\end{center}
\end{figure}

\paragraph{Geometry of style.}
As an insightful application, we consider the geometry of style partially uncovered in this study and produce a low-dimensional representation.
In Fig.~\ref{fig:map}, we propose a \emph{map of style} where we place various oeuvres based on the relative proximity of their corresponding embeddings.
We discuss and interpret this visualization under the lens of literature analysis in Appendix~\ref{app:analysis}.

\paragraph{Interpretability and implications.}
Beyond practicalities, the main objective of this work is to further understand the information content of LLM embeddings.
Many studies have revealed that LLMs construct world models in their latent space, allowing encoding of many features in vector representations, often linearly~\citep{park2024linearrepresentationhypothesisgeometry}.
Some of these features are easily interpretable while others remain obscure~\citep{bricken2023monosemanticity,templeton2024scaling}.
We have shown here that intangible aspects of an input prompt, namely stylistic features, are also abstractly encoded in deep representations.

\section{Conclusion}

We have shown that LLM embeddings representing short ($10^2$ tokens) literary excerpts encode enough information to identify their origin.
Increased confusion between books of the same author
suggests that embeddings convey a stylistic signature specific to a given writer.
This signature appears to lie within a small subspace spanned by the largest principal components.
Shuffling words conserved separability, suggesting the main signal might be about lexical content rather than syntax.

\newpage

\section*{Acknowledgments}
This work was supported by the SciAI Center, and funded by the Office of Naval Research (ONR), under Grant Numbers N00014-23-1-2729 and N00014-23-1-2716.

\bibliography{custom}

\newpage
\appendix

\section{Additional background}
\label{app:background}

\subsection{LLM internal geometry}
LLMs have been found to encode high-level attributes in surprising geometric patterns within their embedding spaces. 
Recent work supports a linear representation hypothesis, wherein certain abstract concepts correspond to linear directions or subspaces in the latent representation space~\citep{park2024linearrepresentationhypothesisgeometry}.
For instance, models appear to linearly represent attributes like factual truthfulness, enabling simple probes or even direct manipulation of activations along those concept directions~\citep{marks2024geometrytruthemergentlinear}.
Similarly, categorical semantic relationships can emerge as geometric structures: models represent categories as vertices of a simplex and encode hierarchical relations via approximately orthogonal components~\cite{park2025geometrycategoricalhierarchicalconcepts}.
The representational geometry induced by prompts has also been analyzed; different prompting or in-context learning strategies imprint distinct geometric signatures on a model's internal states, highlighting how task framing can alter the organization of latent features~\citep{kirsanov2025geometrypromptingunveilingdistinct}.
Moreover, in multilingual settings, LLM embedding spaces can separate language-specific style from language-neutral content along perpendicular axes, suggesting a degree of factorization between surface form and underlying meaning~\citep{chang2022geometrymultilinguallanguagemodel}.

\subsection{Authorship attribution}
An important high-level attribute of interest is literary style. 
Authorship attribution and stylistic analysis have served as tests for whether models capture subtle distributional differences beyond topic or semantics. 
Traditional stylometry relied on carefully engineered linguistic features (e.g. function word frequencies, character n-grams, syntactic patterns), but modern transformer-based LMs can learn such distinctions directly from raw text~\citep{hicke2023t5meetstybaltauthor}.
Recent studies demonstrate that large pretrained models achieve strong performance on author identification.
For example, \citet{hicke2023t5meetstybaltauthor} showed that a fine-tuned T5 model can attribute Early Modern English plays to their likely authors, indicating that LLM representations encode distinctive stylistic signatures. Likewise, GPT-based methods have been applied to Latin prose to verify authorship, with results rivaling traditional stylometric classifiers~\citep{gorovaia-etal-2024-sui}.
Notably, these analyses also highlight that model judgments can be confounded by semantic content rather than pure style.
The challenge of disentangling an author’s unique style from the topic of the text is well-recognized in authorship analysis~\citep{alshomary2024latentspaceinterpretationstylistic}.

\subsection{Interpretability}
\citet{lyu-etal-2023-representation} identified specific latent directions corresponding to concrete stylistic attributes – such as formality and lexical complexity – in a pretrained model’s embedding space. Their findings provide evidence that certain stylistic features are encoded along approximately linear axes, making them separable by simple geometric probes. Such results echo the broader concept-vector findings above, but for attributes of writing style. Still, uncovering literary style (encompassing a complex mix of diction, syntax, and narrative voice) may pose an even greater challenge than these relatively focused style elements.
Interpretability research has begun to bridge these latent representations with human-interpretable descriptions. Some approaches aim to map regions or dimensions of embedding space to understandable concepts. For example, \citet{simhi2023interpreting} project predefined semantic concepts into a model’s embedding space, using them as basis vectors to interpret other embeddings.
Another line of work leverages generative LLMs to produce interpretable style representations: \citet{patel-etal-2023-learning} used GPT-3 to annotate millions of sentences with stylistic descriptors and distilled these into a ``Linguistically Interpretable Style Embedding'' model. The resulting system encodes texts into a 768-dimensional style vector aligned with attributes like formality, tone, and syntax, allowing direct inspection of which dimensions are active for a given text. Together, these efforts underscore both the richness of style information in LLM latent spaces and the complexity of extracting or explaining it.

\section{Methods}
The code for generating and processing data will be made available on a GitHub repository.

\subsection{Dataset}
\label{app:dataset}

In Tab.~\ref{tab:authors}, we present the references to the authors and literary pieces used as the input data to our analysis.

We introduce one text in French (G. Sand's \textit{Jacques}) as a baseline for separability (embeddings from different languages are usually easy to differentiate).
The other texts include different anglophone authors with distinctive styles yet from about the same time period, and from different English-speaking countries. 
This is in order to maintain a certain homogeneity in the style.
We also considered only rather large novels (over 500 pages) in order to be able to assemble substantial ensembles.

\begin{table*}
\centering
\begin{tabular}{c|lccc}
\hline
ID & Author & Novel  & Date & Country\\
\hline
GS & George Sand & \textit{Jacques}$^\star$  & 1833 & France \\
HM1 & Herman Melville & \textit{Moby Dick}  & 1851 & America \\  
HM2 & Herman Melville & \textit{Pierre} & 1852 & America \\ 
JA1 & Jane Austen & \textit{Emma} & 1815 & England \\
JA2 & Jane Austen & \textit{Pride and Prejudice} & 1813 & England \\
JA3 & Jane Austen & \textit{Sense and Sensibility} & 1811 & England \\
JJ & James Joyce & \textit{Ulysses} & 1922 & Ireland \\
MT1 & Mark Twain & \textit{Life on the Mississippi} & 1883 & America\\
MT2 & Mark Twain & \textit{Roughing It} & 1872 & America \\
NH1 & Nathaniel Hawthorne & \textit{The House of the Seven Gables} & 1851 & America \\
NH2 & Nathaniel Hawthorne & \textit{The Scarlet Letter} & 1850 & America \\
VW1 & Virginia Woolf & \textit{Night and Day} & 1919 & England \\
VW2 & Virginia Woolf & \textit{The Voyage Out} & 1915 & England \\
\hline
\end{tabular}
\caption{
Authors and novels used for analysis. Note that \textit{Jacques} is in French.
}
\label{tab:authors}
\end{table*}

\subsection{Excerpts}
The full text of a given novel is tokenized at once, and the resulting sequence of token identifiers is divided into chunks of $N$ consecutive tokens.
These non-overlapping chunks represent short text fragments of various forms. 
In particular, they do not necessarily start or end with a sentence and the last token can be anything: punctuation, word suffix, article, verb, etc.
In Tab.~\ref{tab:excerpts}, we show a few examples of 16-token chunks in their textual form.

\begin{table*}
\centering
\begin{tabular}{r|l}
\hline
15 preceding tokens & last token \\
\hline
\small \verb|least knew somebody who knew his father and mother? To the peasants of old | & \small \verb|time |\\
\small \verb|off time superstition clung easily round every person or thing that was at all | & \small \verb|unw | \\
\small \verb|crime; especially if he had any reputation for knowledge, or showed any skill | & \small \verb|in |\\
\small \verb|live in a rollicking fashion, and keep a jolly Christmas, | & \small \verb|Wh| \\
\small \verb|shook him, and his limbs were stiff, and his hands clutched the | & \small \verb|bag | \\ 
\small \verb|nothing strange for people of average culture and experience, but for the villagers near | & \small \verb|whom |\\
\small \verb|road, and lifting more imposing fronts than the rectory, which | & \small \verb|peeped| \\
\small \verb|which seemed to explain things otherwise incredible; but the argumentative Mr. | & \small \verb|Macey| \\
\small \verb|handicraft. All cleverness, whether in the rapid use of that difficult | & \small \verb|instrument| \\
\small \verb|lar or the knife-grinder. No one knew where wandering men had | & \small \verb|their| \\
\small \verb|-weaver, named Silas Marner, worked at his vocation in | & \small \verb|a| \\
\small \verb|of it, and two or three large brick-and-stone homesteads| & \small \verb|, | \\
\small \verb|the outskirts of civilization—inhabited by meagre sheep and thinly-| & \small \verb|sc| \\
\small \verb|certain awe at the mysterious action of the loom, by a pleasant sense of | & \small \verb|scorn | \\
\small \verb|The questionable sound of Silas’s loom, so unlike the natural cheerful | & \small \verb|tro| \\
\hline
\end{tabular}
\caption{
Examples of 16-token excerpts from G. Eliot's \textit{Silas Marner}.
Embeddings derived from the last token are the ones collected to form the ensembles of the analysis.
}
\label{tab:excerpts}
\end{table*}

\subsection{Classifiers}
For binary classification, we use a Support Vector Machine with linear kernel, and after dimensionality reduction by PCA to 64 dimensions. 
The training to validation data split was 70/30.

For multiclass classification, we train a Multilayer Perceptron with penultimate layer of dimension 32, cross-entropy loss and Adam optimizer.

\section{Map of style}
\label{app:analysis}

Fig.~\ref{fig:map} presents a visualization of style proximity across books and authors. 
Here we propose an alternative representation and interpret it from a traditional literature analysis perspective.

\subsection{Centroid visualization}
\begin{figure}[htbp]
\begin{center}
\centerline{\includegraphics[width=\columnwidth]{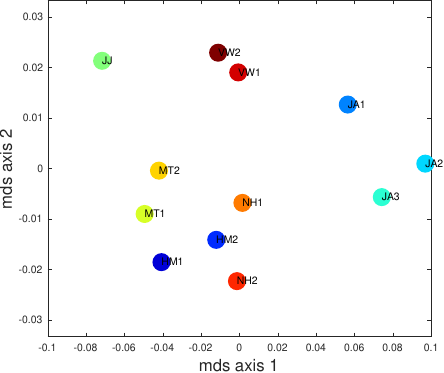}}
\caption{
Alternative map of style emphasizing relative distances between ensembles by embedding their shared geometry with multidimensional scaling.
}
\label{fig:centroid}
\vspace{-2em}
\end{center}
\end{figure}

In order to emphasize proximity between ensembles (rather than text snippets), we propose an alternative representation based on centroid proximity.
We calculate the centroids location of each book ensemble and apply multidimensional scaling (MDS) to yield a two-dimensional representation in Fig.~\ref{fig:centroid}.
The similarity matrix used for MDS is the pair cosine distances between centroids.

\subsection{Literary comment}
This spatial distribution of Fig.~\ref{fig:centroid} depicts interesting relationships between the narrative techniques and styles of various authors. Virginia Woolf (VW) and James Joyce (JJ), known for pioneering techniques like stream of consciousness and free indirect discourse, cluster apart, reflecting their shared modernist experimentation. Jane Austen’s texts (JA), which also employ free indirect discourse but within a more traditional realist framework, form their own distinct grouping. Austen's group is clearly separated both from the modernists and from American authors such as Nathaniel Hawthorne (NH), Herman Melville (HM), and Mark Twain (MT). These American writers, characterized predominantly by narrative realism or romanticism, are grouped centrally and distinctly apart from the experimental modernist approaches of Woolf and Joyce.

\section{Generalization to additional models}
\label{app:models}

For generalization, we reproduce the same methodology with three other open-source models released in 2025: 
\begin{itemize}
\item \verb|gemma-3-1b-pt| from Google (US)~\citep{gemmateam2025gemma3technicalreport}
\item \verb|Qwen3-1.7B-Base| from Qwen (China)~\citep{yang2025qwen3technicalreport}
\item \verb|SmolLM2-1.7B| from Hugging Face (France)~\citep{allal2025smollm2smolgoesbig}
\end{itemize}
These models are queried in their ``base'' form, i.e. not fine-tuned for chat (``instruct'').
We use models in the one-billion-parameter range as smaller models are generally more interpretable, and also due to compute constraints.

\begin{figure}[htbp]
\begin{center}
\centerline{\includegraphics[width=\columnwidth]{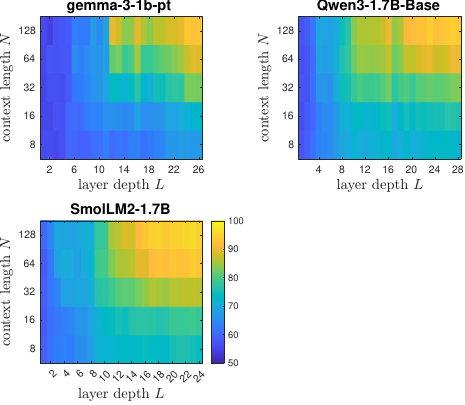}}
\caption{
Probe accuracy (\%) for GE vs VW in other LLMs (applied on deepest embeddings).
}
\label{fig:gemma}
\vspace{-2em}
\end{center}
\end{figure}

We find the same patterns of ensemble separability across models, as shown in Fig.~\ref{fig:gemma}.
In particular, embeddings corresponding to increased context and deep layers are more easily separable.

\end{document}